\documentclass[final]{cvpr}

\usepackage{times}
\usepackage{epsfig}
\usepackage{graphicx}
\usepackage{amsmath}
\usepackage{amssymb}
\usepackage{multirow}
\usepackage{makecell}
\usepackage{url}
\usepackage[numbers, sort]{natbib}
\usepackage{verbatim}
\usepackage{algorithm}
\usepackage{algorithmicx}
\usepackage{algpseudocode}
\usepackage{makecell}
\usepackage{textcomp}
\usepackage{subcaption}
\usepackage{amsthm,amsmath,amssymb}
\usepackage{mathrsfs}
\usepackage{comment}
\usepackage{bbding}
\usepackage{comment}
\usepackage{color}
\usepackage{bm}
\usepackage[pagebackref=true,breaklinks=true,colorlinks,bookmarks=false]{hyperref}

\def\cvprPaperID{685} % *** Enter the CVPR Paper ID here
\def\confYear{CVPR 2021}

\begin{document}

%%%%%%%%% TITLE
\title{Structure-Aware Face Clustering on a Large-Scale Graph with $\bf{10^{7}}$ Nodes}

\author{Shuai Shen\textsuperscript{1,2}, Wanhua Li\textsuperscript{1,2}, Zheng Zhu\textsuperscript{1,2}, Guan Huang\textsuperscript{3}, Dalong Du\textsuperscript{3}, Jiwen Lu\textsuperscript{1,2,}\thanks{\;Corresponding author}\ ,  Jie Zhou\textsuperscript{1,2} \\
\textsuperscript{1} Department of Automation, Tsinghua University, China\\
\textsuperscript{2} Beijing National Research Center for Information Science and Technology, China\\
\textsuperscript{3} XForwardAI\\
	{\tt \small \{shens19,li-wh17\}@mails.tsinghua.edu.cn;zhengzhu@tsinghua.edu.cn;}\\
	{\tt \small \{guan.huang,dalong.du\}@xforwardai.com;\tt \small \{lujiwen,jzhou\}@tsinghua.edu.cn}}

\maketitle
\pagestyle{empty}  % no page number for the second and the later pages
\thispagestyle{empty} % no page number for the first page

%%%%%%%%% ABSTRACT
\begin{abstract}
Face clustering is a promising method for annotating unlabeled face images. Recent supervised approaches have boosted the face clustering accuracy greatly, however their performance is still far from satisfactory. These methods can be roughly divided into global-based and local-based ones. Global-based methods suffer from the limitation of training data scale, while local-based ones are difficult to grasp the whole graph structure information and usually take a long time for inference. Previous approaches fail to tackle these two challenges simultaneously. To address the dilemma of large-scale training and efficient inference, we propose the STructure-AwaRe Face Clustering (STAR-FC) method. Specifically, we design a structure-preserved subgraph sampling strategy to explore the power of large-scale training data, which can increase the training data scale from ${10^{5}}$ to ${10^{7}}$. During inference, the STAR-FC performs efficient full-graph clustering with two steps: graph parsing and graph refinement. And the concept of node intimacy is introduced in the second step to mine the local structural information. The STAR-FC gets 91.97 pairwise F-score on partial MS1M within 310s which surpasses the state-of-the-arts. Furthermore, we are the first to train on very large-scale graph with 20M nodes, and achieve superior inference results on 12M testing data. Overall, as a simple and effective method, the proposed STAR-FC provides a strong baseline for large-scale face clustering. Code is available at \url{https://sstzal.github.io/STAR-FC/}.
\end{abstract}
%%%%%%%%% BODY TEXT
%-------------------------------------------------------------------------
\section{Introduction}
Recent years have witnessed the great progress of face recognition~\cite{sun2014deep,taigman2014deepface,liu2016large,wang2018cosface,liu2017sphereface,deng2019arcface}. Large-scale datasets are an important factor in the success of face recognition and there is an increasing demand for larger-scale data. Face clustering~\cite{lloyd1982least,zhan2018consensus,wang2019linkage,yang2020learn,zhan2020online,li2020deep,zhang2020global} is a natural way to solve the data annotation problem so as to make better use of massive unlabeled data. Face clustering is also one possible approach to organize and file large volumes of real face images in social media or other application scenarios.

\begin{figure}
\begin{center}
\includegraphics[width=0.94\linewidth]{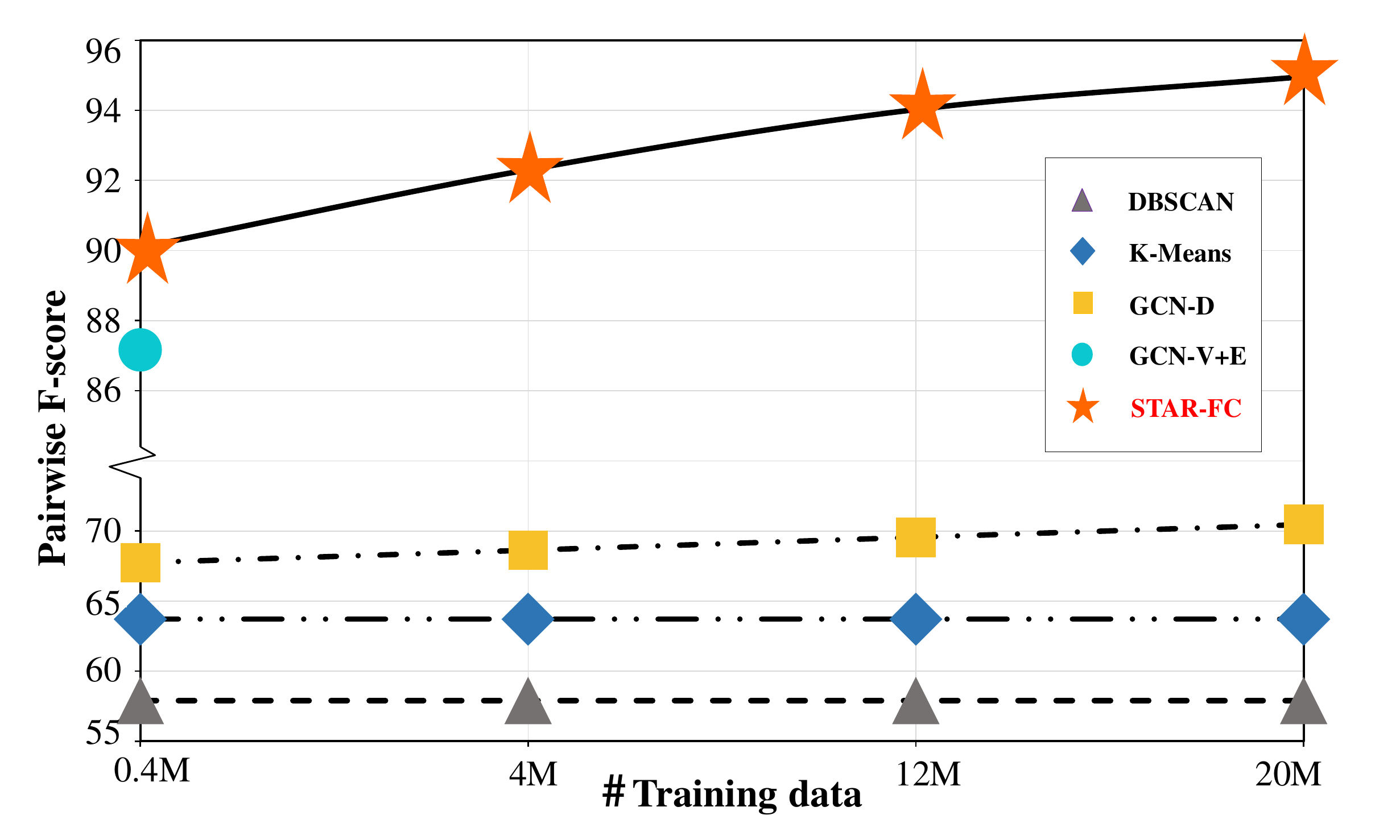}
\end{center}
\vspace{-8mm}
\caption{Method comparison when training with different scales of data and testing on 12M data from WebFace42M~\cite{FB}. The proposed STAR-FC can fully explore the power of large-scale training data. GCN-V+E fails to handle larger training graph while GCN-D's performance is severely restricted due to the less consideration of the global structural information.}
\vspace{-6mm}
\label{fig1}
\end{figure}

Recently a variety of efforts have been devoted to face clustering. Traditional unsupervised methods~\cite{ho2003clustering,zhao2006automatic} including K-Means~\cite{lloyd1982least} and DBSCAN~\cite{ester1996density} usually depend on some manually designed clustering strategies. They perform well on small datasets, however they are less effective when dealing with large-scale data as shown in Figure~\ref{fig1}. Recent research trends~\cite{zhan2018consensus,wang2019linkage,yang2019learning,yang2020learning,guo2020density} turn to the GCN-based supervised learning. These methods are performed based on the affinity graph and can be roughly divided into global-based and local-based ones \emph{according to whether their GCN input is the whole graph or not}. The representative global-based method GCN-V+E~\cite{yang2020learning} uses the entire graph for GCN training. As shown in Figure~\ref{fig1}, it boosts the face clustering performance greatly compared with unsupervised methods, however the training data scale is limited by the GPU memory which makes it difficult to further explore the power of larger-scale training sets. Although local-based methods such as GCN-D~\cite{yang2019learning} shown in Figure~\ref{fig1} don't suffer from memory restrictions, its performance is limited since it lacks the comprehension of global graph structure. Besides, it organizes the data as overlapped local subgraphs which leads to inefficient inference.

%and it affects the performance upper bound to a large extent. 
For many computer vision tasks~\cite{lin2014microsoft,deng2009imagenet,ren2015faster,girshick2015fast,he2016deep,deng2020sub}, large-scale training data is one of the most important engines to promote the performance. With the emergence of some new large-scale benchmarks such as WebFace42M~\cite{FB} whose data volume is ten times that of MS1M~\cite{guo2016ms}, we have more data available for training. Therefore, exploring the power of these rich training data is imperative. For testing, efficiency matters, we are therefore eager to perform full graph inference. Based on the above motivations, we propose a structure-aware face clustering method STAR-FC to address the dilemma of large-scale training and efficient inference. Specifically, we design a GCN~\cite{kipf2016semi} based on the $K$NN~\cite{cover1967nearest} affinity graph to estimate the edge confidence. Furthermore, a structure preserved subgraph sampling strategy is proposed for larger-scale GCN training. During inference, we perform face clustering with two steps: graph parsing and graph refinement. In the second step, node intimacy is introduced to mine the local structural information for further refinement. In the inference process, the entire graph is taken as the input for efficiency.

%thus break through the original performance upper bound.

%All above operations are implemented with full graph operation and parallel matrix computation leading to efficient inference.

%We can therefore gradually distill face clusters from the affinity graph.

%Compared with full-graph training, such sampling strategy can avoid performance loss and bring further gains since it can enhance the generalization of the model. 

%the affinity graph is firstly parsed with the learned GCN. However, there still exist a minority of false posotive edges hard to be removed through edge confidence. We therefore introduce the concept of node intimacy to mine the local structural information for further refinement. 

%The experiments demonstrate that the STAR-FC can not only handle large-scale training on 20M data but also perform efficient inference. It achieves state-of-the-art performance on face clustering and improves the accuracy of face recognition to be comparable with the supervised method. 

The experiments demonstrate that with these structure-aware designs, the STAR-FC can not only perform sample-based training but also implement full-graph inference. With sample-based training, the training data scale can be increased by two orders of magnitude from $10^{5}$ to $10^{7}$ and beyond. As shown in Figure~\ref{fig1}, with the increase of training data, our method has been consistently improved and finally achieves 95.1 \emph{pairwise F-score}. Interestingly, we find that such sampling method does not lead to performance loss and brings some extra accuracy gain since it enhances the generalization of the
model. In inference, with full-graph as input, the efficiency can be guaranteed. We achieve state-of-the-art face clustering results on partial MS1M within 310s. What's more, we can complete inference on 12M data within 1.7h thus provide a  strong baseline for face clustering.
%It achieves state-of-the-art performance on face clustering and improves the accuracy of face recognition to be comparable with the supervised method. 
To summarize, we make the following contributions:
\vspace{-5mm}
\begin{itemize}
\item To fully explore the power of large-scale training dataset, we propose a structure-preserved subgraph sampling strategy which can break through the limitation of training data scale from $10^{5}$ to $10^{7}$.
\vspace{-2mm}
\item For inference, we take the entire graph as input to ensure efficiency. We transform face clustering into two steps: graph parsing and graph refinement. Node intimacy is introduced in the second step to explore the local structure for further graph refinement.

%introduce the concept of node intimacy for face clustering. It aims to explore the local graph structure through neighbor information.
\vspace{-2mm}
\item The proposed STAR-FC achieves 91.97 \emph{F-score} on partial MS1M within 310s. Moreover, we are the first to conduct large-scale training on 20M data which provides a strong baseline for large-scale face clustering. 
\end{itemize}

%-------------------------------------------------------------------------
\section{Related Work}

\noindent\textbf{Face Clustering}. 
Face clustering has been extensively studied as a classic task in machine learning. It provides an alternative way to exploit massive unlabeled data. Traditional algorithms including K-Means~\cite{lloyd1982least}, spectral clustering~\cite{ho2003clustering}, hierarchical clustering~\cite{zhao2006automatic} and DBSCAN~\cite{ester1996density} laid a good theoretical foundation for clustering. However, they generally rely on simple data distribution assumptions thus are ineffective when dealing with real data. To improve the robustness in complex distributed face clustering, Lin~\emph{et al.}~\cite{lin2017proximity} proposed the proximity-aware hierarchical clustering. Zhu~\emph{et al.}~\cite{zhu2011rank} and Otto~\emph{et al.}~\cite{otto2017clustering} designed the rank-order connection metric. However, since ~\cite{zhu2011rank,otto2017clustering} did not establish the graph structure and lacked preliminary analysis for neighbor relations, they achieved poor results. Lin~\emph{et al.}~\cite{lin2018deep} tried to measure density affinities between local neighborhoods.~\cite{shi2018face} modeled face clustering as a structured prediction problem using conditional random field.
%Otto et al.~\cite{otto2017clustering} further improved~\cite{zhu2011rank} to the approximate rank-order metric.

Methods above make less use of supervised information in face clustering. More recently, the research trends turn to GCN-based supervised face clustering and have achieved impressive results~\cite{zhan2018consensus,wang2019linkage,yang2019learning,yang2020learning,guo2020density,zhang2020global,yang2020learn}. These methods can be roughly divided into two categories: local-based face clustering~\cite{zhan2018consensus,wang2019linkage,yang2019learning} and global-based one~\cite{yang2020learning}. In these local-based methods, Zhan~\emph{et al.}~\cite{zhan2018consensus} designed a mediator network to aggregate information in the local graph. Wang~\emph{et al.}~\cite{wang2019linkage} predicted the linkage in an instance pivot subgraph. Yang~\emph{et al.}~\cite{yang2019learning} generated a series of subgraphs as proposals and detected face clusters thereon. These methods pay more attention to local graph information and rely heavily on redundancy subgraph operations which limit their performance and lead to slow inference. Representative global-based method~\cite{yang2020learning} took the entire graph as input and predicted the confidence and connectivity of all vertices. In~\cite{yang2020learning}, the holistic graph structure is better considered, however due to GPU memory limitation, it may be out of memory when dealing with larger training data. We therefore propose the STAR-FC to simultaneously tackle the challenges of large-scale training and efficient inference.

\begin{figure*}[tb]
\begin{center}
\includegraphics[width=0.94\linewidth]{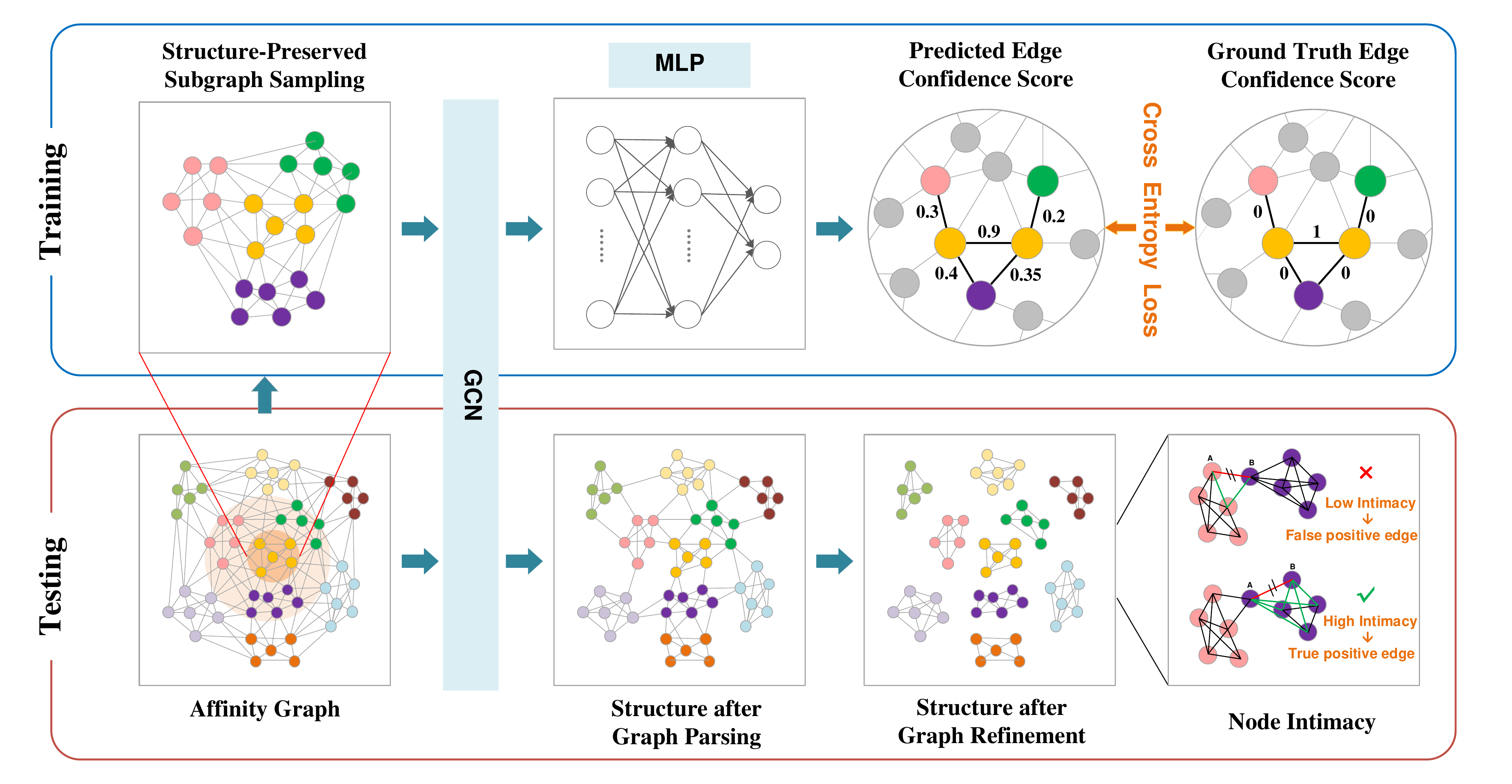}
\end{center}
\vspace{-6mm}
\caption{Overview of the proposed STAR-FC framework. In the training process, we use the structure-preserved subgraph sampling strategy to get a variety of subgraphs which are used to train the GCN-based edge confidence estimator. The cross-entropy loss is employed to supervise the training. During inference, based on the built affinity graph, we transform the face clustering into two steps: graph parsing and graph refinement. For the first step, the trained GCN takes the entire graph as input and estimates all edge confidence scores simultaneously. The affinity graph is parsed with these predicted scores. In the second step, node intimacy is employed for further graph refinement. After these two steps, the cluster structure will become clear and the face clusters can be directly read from the graph. For details about the node intimacy, please refer to Figure~\ref{NI}.}
\vspace{-4mm}
\label{overview}
\end{figure*}

\vspace{4pt}
\noindent\textbf{Graph Convolutional Networks}. Graph Convolutional Networks (GCNs)~\cite{wu2020comprehensive,schlichtkrull2018modeling,velivckovic2017graph,chiang2019cluster} extend the convolution idea of CNNs~\cite{chollet2017xception,krizhevsky2017imagenet,parkhi2015deep} to process non-Euclidean structured data. GCNs have shown impressive capability on various tasks~\cite{yan2018spatial,yan2019convolutional,berg2017graph,kipf2016semi,li2020grapheccv,li2020graphicme}. More recently, to improve GCN's ability in handling larger-scale training graph, some GCN training algorithms have been proposed. GraphSAGE~\cite{hamilton2017inductive} traded off the performance and runtime via sampling a fixed number of neighbors for aggregation. FastGCN~\cite{chen2018fastgcn} interpreted graph convolutions as integral transforms of embedding functions under probability measures and proposed to sample vertices rather than neighbors for controllable computation cost. Whereas the key differences between the above methods and the proposed one lie in the sample mode. These previous methods perform graph sampling with nodes as the smallest unit, while the proposed method implements sampling on clusters with near neighbor relationships trying to approximate global structure. Our method can keep most of the inter-cluster edges which can be provided as hard negative samples during training. 

%-------------------------------------------------------------------------

\section{Methodology}

\subsection{Overview}
\label{over}
%Supervised approaches have shown their superiority in face clustering with complex cluster patterns. Nevertheless, their performance is still far from satisfactory with the limitations on training data scale or local graph operations corresponding to the global-based and local-based methods. 
To address the dilemma of large-scale training and efficient inference, we propose a structure-aware face clustering method. An overview of the proposed STAR-FC is shown in Figure~\ref{overview}. During training, the GCN-based edge confidence estimator is trained with the structure-preserved subgraph sampling strategy. We aim to approximate the full graph structure with the sampled subgraph and it retains most of the hard negative edges which contribute a lot for training. In this way, the potential of large-scale data can be fully unleashed. We specifically model the edge prediction as a binary classification problem and use the cross-entropy loss for supervising. During inference, we transform the face clustering into two steps: graph parsing and graph refinement. For graph parsing, we take the whole graph as the input of the trained GCN to estimate all edge confidence scores simultaneously. Then edges with low scores are removed and graph structures will become clearer. However there still exist a few wrong connections. They have relatively high scores thus hard to be eliminated. To further refine the graph, we introduce the node intimacy for edge pruning again. After these two steps, face clusters are naturally formed by those connecting groups in the graph.

\subsection{Large-Scale GCN Training}
\label{train}
In this section, we detail the large-scale training process of the proposed STAR-FC.

\vspace{3pt}
\noindent\textbf{Design of the GCN}. In this step, we design a GCN-based edge confidence predictor on the basis of $K$NN affinity graph. We first get the feature matrix $F\in \mathbb{R}^{N\times D}$ with a trained ResNet-50, where $N$ is the number of face images and $D$ is the feature dimension. To build the $K$NN affinity graph, each sample can be treated as a node in the graph and is linked to its $K$ nearest neighbours~\cite{cover1967nearest}. The corresponding sparse symmetric adjacency matrix is $A\in \mathbb{R}^{N\times N}$. 

Since the CNN is trained under the strong supervision of classification loss, the extracted features $F$ actually contain rich identity information. However, due to intra-class variance and the fixed $K$ value in $K$NN algorithm, the affinity graph may contain many wrong edge connections. We therefore try to directly predict the existence of the edges employing a GCN for propagating neighbor information.
Following~\cite{wang2019linkage, yang2020learning}, we use the more effective modified GCN as our backbone and the computational process of a $L$-layer modified GCN can be formulated as:
\begin{equation} 
F_{l+1}=\sigma \left ( [F_{l}^{T},( \widetilde{A}F_{l} ) ^{T}]^{T} W_l \right ), \label{softmax}
\end{equation}
where $ \widetilde{A}=\widetilde{D}^{-1}\left ( A+I \right )$. $\widetilde{D}$ is a diagonal degree matrix with $\widetilde{D}=\sum_{j}\widetilde{A}_{ij}$. $F_{l}$ denotes the embeddings at $l$-th layer and $F_{0}$ is the input face features. $W_{l}\in \mathbb{R}^{D_{in}\times D_{out}}$ is a learnable matrix that maps the embeddings to a new space. $\sigma$ is a nonlinear activation and we use ReLU~\cite{xu2015empirical,nair2010rectified} in this work. $F_{L}$ indicates the output features of $L$-layer.

%by connecting the input embeddings and embeddings after neighborhood aggregation in each layer. Specifically, given the adjacency matrix $A$ and the feature matrix $F$, the computational process of a $L$-layer residual GCN can be formulated as:
%\begin{equation} 
%F_{l+1}=\sigma \left ( \widetilde{A}F_{l} W_l \right ), \label{softmax}
%\end{equation}

%\begin{equation} 
%F_{l+1}=\sigma \left ( [F_{l}^{T},( \widetilde{A}F_{l} ) ^{T}]^{T} W_l \right ). \label{softmax}
%\end{equation}
%where $ \widetilde{A}=\widetilde{D}^{-1}\left ( A+I \right )$ and $\widetilde{D}$ is a diagonal degree matrix with $\widetilde{D}=\sum_{j}\widetilde{A}_{ij}$.
%$F_{l}$ denotes the embeddings at $l$-th layer and $F_{0}$ is the input face features. $W_{l}\in \mathbb{R}^{D_{in}\times D_{out}}$ is a learnable matrix that maps the embeddings to a new space. $\sigma$ is a nonlinear activation and we use ReLU in this work. 

Since $F_{L}$ aggregates many messages from the neighborhood and encode graph structural information, it is more suitable for the face clustering task. To predict the existence of the edges in the affinity graph, we design a binary classifier employing a 2-layer MLP~\cite{jain1996artificial} with the objective to minimize the cross-entropy loss between the predicted edge confidence and the ground-truth edge labels. Particularly, we take pair features corresponding to the edges in the affinity graph as the input of the MLP and get the two-dimension predicted edge confidence. The ground-truth label of an edge is 1 if the two nodes connected by this edge belong to the same class, otherwise it will be 0.

Under this simple supervision of binary signals, the distribution of the predicted confidence will appear as two sharp peaks approaching 0 and 1 as shown in Figure~\ref{distri}. Therefore, for inference we can use a single threshold $\tau_{1}$ to effectively eliminate most of the wrong edges. This operation may lead to two types of misjudgments: (1) it may cut off a small number of real edges; (2) it may leave a few wrong connections hard to be identified via confidence. Since the original affinity graph is densely connected, the lost correct edges in the former case have a slight effect on the connectivity of the final graph. Those remaining wrong edges will be processed in the following procedure with node intimacy in Section~\ref{local}.

\vspace{3pt}
\noindent\textbf{Structure-Preserved Subgraph Sampling.} Previous methods~\cite{yang2019learning,yang2020learning} usually use 10\% of MS1M (0.5M face images)~\cite{guo2016ms} for GCN training. For global-based method~\cite{yang2020learning} this has approximated to the memory threshold of a typical 1080Ti GPU. Although local-based methods~\cite{yang2019learning,wang2019linkage} can alleviate the GPU storage burden through local graph operation, they rely heavily on numerous overlapped subgraphs which severely affect their efficiency and accuracy.

\begin{algorithm}[t]
 \caption{Structure-Preserved Subgraph Sampling}
 \begin{algorithmic}[1]
 \renewcommand{\algorithmicrequire}{\textbf{Input:}}
 \renewcommand{\algorithmicensure}{\textbf{Output:}}
 \Require Training nodes reorganized in clusters $C$, cluster seeds number $M$, parameters $N$. 
 \Ensure  Sampled subgraph $\mathcal{S}$
 \State  $\mathcal{S}=\emptyset$, $\mathcal{S}_1=\emptyset$, $\mathcal{S}_2=\emptyset$
  \State Randomly select $M$ clusters $\mathcal{C}_{i}$ ($i=1,\cdots, M$) from $C$
  %\State $\mathcal{S}_1 = \mathcal{S}_1 \bigcup_{i=1}^{M} \mathcal{C}_{i}$
  \For{$i=1$ to $M$}
   %\State $\mathcal{S}_1=\emptyset$
     \State \parbox[t]{0.9\linewidth} {Sample $N$ nearest neighbor clusters $\mathcal{C}_{ij}$ ($j=1,\cdots, N$) of $\mathcal{C}_{i}$}
     \State $\mathcal{S}_1 = \mathcal{S}_1 \bigcup \mathcal{C}_{i} \bigcup_{j=1}^{N} \mathcal{C}_{ij}$
       \vspace{1mm}
       %\State \parbox[t]{0.9\linewidth}{Construct $\mathcal{S}_2$ via randomly selecting $K_1$ clusters from $\mathcal{S}_{1}$ (R1)}
       %\State \parbox[t]{0.9\linewidth}{Construct $\mathcal{S}_3$ via randomly sampling $K_2$ percent nodes from each cluster in $\mathcal{S}_2$ (R2)}
     % \State  $\mathcal{S}_2$ =  $\mathcal{S}_2 \bigcup \mathcal{S}_1$ 
  \EndFor
          \State \parbox[t]{0.9\linewidth}{Construct $\mathcal{S}_2$ via applying \emph{CR} on $\mathcal{S}_{1}$}
       \State \parbox[t]{0.9\linewidth}{Construct $\mathcal{S}$ via applying \emph{SR} on $\mathcal{S}_2$}
\State \Return $\mathcal{S}$
 \end{algorithmic}
 \label{alg:sample}
\end{algorithm}

%We make a statistics about the structure of the KNN graph of 10\% MS1M with $K=80$, results show that nodes within the same cluster are densely connected by about 70\% edges, while there exists a few sparse connections (accounts for 30\% of the total edges) between near clusters. 

Recent years have witnessed the success of large-scale training in many computer vision tasks~\cite{lin2014microsoft,deng2009imagenet,ren2015faster,girshick2015fast,he2016deep,deng2020sub}. To fully explore the power of large-scale training data, we design a structure-preserved subgraph sampling (SPSS) strategy for GCN training. The edges in an affinity graph are mainly composed of two parts: dense intra-cluster connections and sparse connections between near clusters. Try to approximate the dense connections within clusters, our approach treats face clusters as the smallest sampling unit, different from previous methods~\cite{chen2018fastgcn,hamilton2017inductive} which perform randomly sampling on nodes. To further model those inter-cluster connections, we extend the subgraph from the selected cluster to its neighbor clusters.
On one hand, the sampled subgraphs preserve the important structural information of the full graph, \ie the edge connections within the clusters and the connections between near clusters. On the other hand, many near clusters are sampled in a subgraph, and the edges between these near clusters can be treated as hard negative  examples which can contribute a lot to the GCN training. Equipped with such a structure-preserved subgraph sampling strategy, our method can benefit from increasing training data. Interestingly, this sampling strategy does not lead to performance loss since the structural information of the whole graph is fully considered. Besides, the experimetal results in Table~\ref{table1} show that it brings further performance gain since the enhancement of generalization.

Algorithm~\ref{alg:sample} shows the details of the proposed SPSS. Given the training nodes reorganized in clusters, we randomly select $M$ clusters from them as the sampling seeds. For each seed cluster, we extend to its $N$ nearest neighbor clusters which are measured by the cosine similarity of the center. After this step, we can get a subgraph $\mathcal{S}_1$ consists of $M\times N$ clusters. To further strengthen the generalization, we introduce the \emph{cluster randomness} (CR) strategy by randomly selecting $K_1$ clusters from $\mathcal{S}_1$, and the \emph{sample randomness} (SR) strategy by randomly selecting $K_2$ nodes from $\mathcal{S}_2$. Then we rebuild the \emph{K}NN affinity graph based on these sampled nodes to construct the subgraph $\mathcal{S}$.
%In this way, similar nodes can be sampled in a subgraph which can bring more valuable neighbor message for the GCN aggregation operation.

%Besides, it enhances the generalization of the trained model to some extent and bring further performance gain as shown in Table~\ref{table1}.

\subsection{Efficient Face Clustering Inference}
\label{local}
This subsection shows how to perform efficient inference and get the final face clusters with the proposed STAR-FC in detail. Specifically, we transform the face clustering task into two steps: graph parsing and graph refinement.

\begin{figure}
\begin{center}
\includegraphics[width=0.8\linewidth]{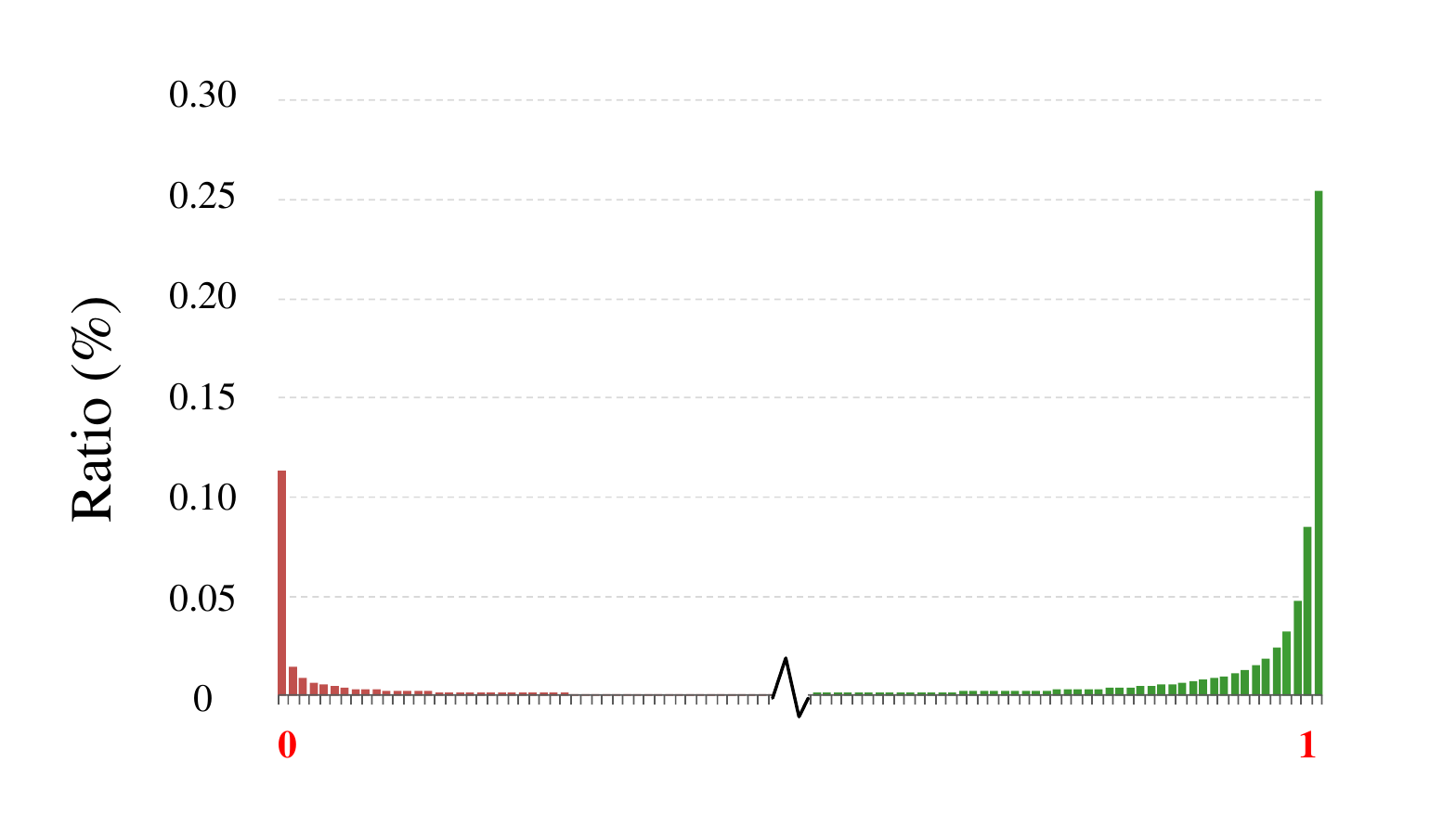}
\end{center}
\vspace{-10mm}
\caption{Distribution of the predicted edge scores obtained by the trained GCN. Since the GCN is trained under the supervision of binary signals, its predicted scores appear as two sharp peaks approaching 0 and 1 with high separability.}
\vspace{-4mm}
\label{distri}
\end{figure}

\vspace{2pt}
\noindent\textbf{Graph Parsing.} In this step, we parse the built affinity graph preliminarily with the trained GCN. We feed the entire graph into the GCN and obtain all edge scores simultaneously. Figure~\ref{distri} shows that the predicted scores are distributed in nearly two sharp peaks approaching 0 and 1. We can therefore perform simple but effective pruning with a single threshold $\tau_{1}$. A small number of correct edges may be cut off wrongly in this step. Whereas since the initial affinity graph is densely connected, this has a slight effect on the connectivity of the final clusters. After this step, most wrong connections are removed, and the graph structure has become clearer. However, there still exists a minority of false positive edges. To deal with these edges, we propose node intimacy in the second step trying to mine the local graph structure for further graph refinement.

\begin{comment}
\begin{algorithm}[t]
 \caption{Inference}
 \begin{algorithmic}[1]
 \renewcommand{\algorithmicrequire}{\textbf{Input:}}
 \renewcommand{\algorithmicensure}{\textbf{Output:}}
 \Require Affinity Graph $\mathcal{G}$, Trained GCN $G$, thresholds $\tau_{1}$ and $\tau_{2}$ . 
 \Ensure  Face clusters $C$
 \State  Estimate all edge scores in $\mathcal{G}$ via $G$
 \State  Remove edges whose score is below $\tau_{1}$ and get the refined graph ${\mathcal{G}}'$
 \State  Remove edges whose score is below $\tau_{1}$
 \State Calculate NI of all linked nodes in ${\mathcal{G}}'$
 \State  Remove edges whose corresponding NI is below $\tau_{2}$ and get the refined graph ${\mathcal{G}}''$
 \State Read face clusters $C$ from ${\mathcal{G}}'$
\State \Return $\mathcal{S}$
 \end{algorithmic}
 \label{alg:sample}
\end{algorithm}
\end{comment}

\vspace{2pt}
\noindent\textbf{Graph Refinement.} The left false positive edges mistakenly connect different clusters, which may seriously affect the final clustering performance. These edges can not be removed directly using edge scores, we therefore try to identify them with node intimacy (NI).

\begin{figure}
\begin{center}
\includegraphics[width=0.9\linewidth]{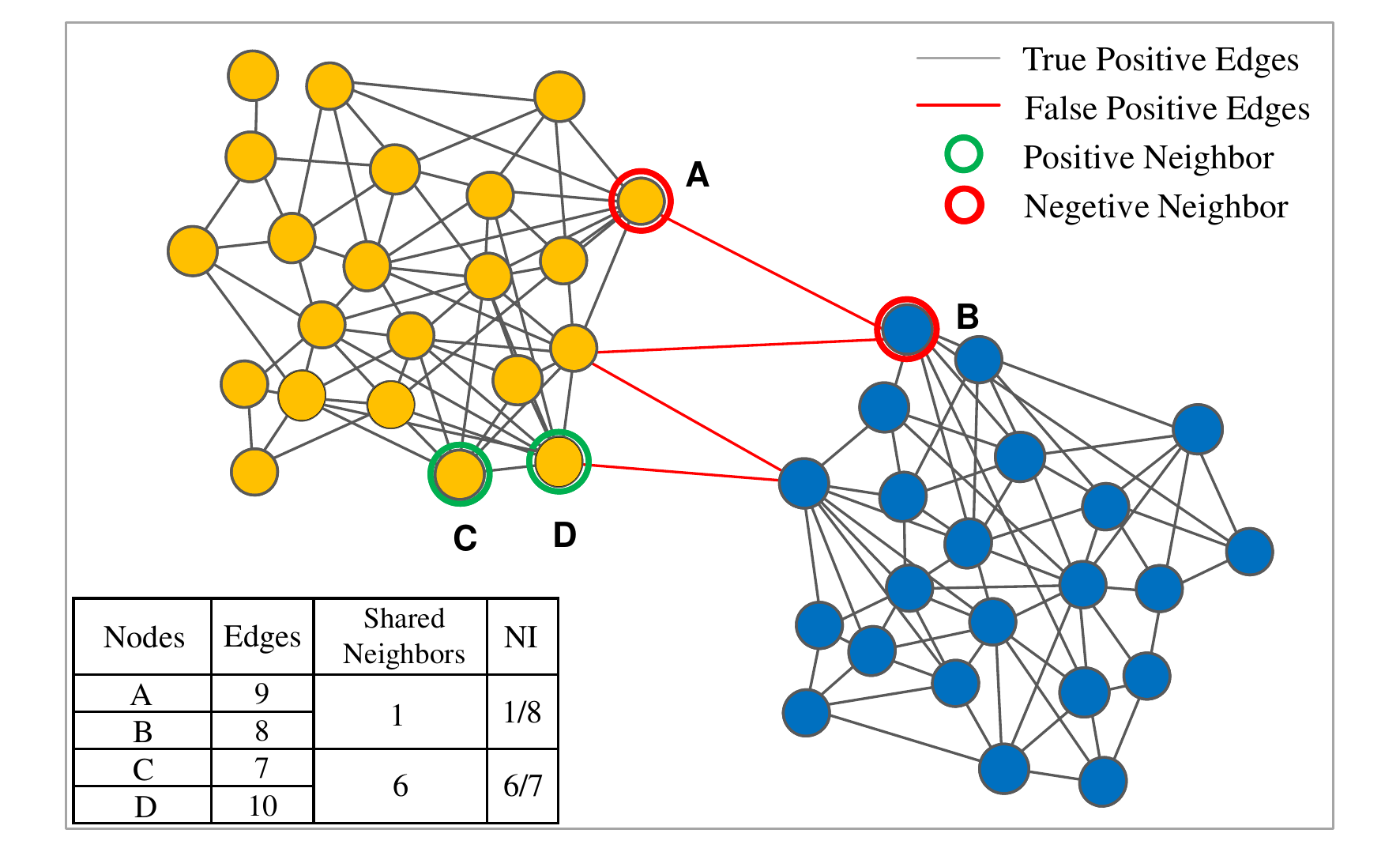}
\end{center}
\vspace{-7mm}
\caption{Illustration of the proposed node intimacy. We observe that nodes within a cluster are densely connected with each other while there exist sparse wrong connections between clusters. Inspired by such observation, we present the node intimacy which aims to estimate whether two nodes should be linked by measuring their common neighbors. The table at the bottom left shows the specific NI calculation of two pairs of nodes (\ie A\&B, C\&D). The results indicate that the positive neighbor C\&D exactly have high NI while the negative neighbor A\&B have low NI.}
\vspace{-5mm}
\label{NI}
\end{figure}

The concept of node intimacy is inspired by the human familiarity. In human societies, two familiar people usually have many mutual friends. Extending this idea to the graph, we establish the concept of node intimacy. Particularly, given two nodes $\mathcal{N}_{1}$ and $\mathcal{N}_{2}$. There are $n_{1}$ edges connected to $\mathcal{N}_{1}$ while $n_{2}$ edges connected to $\mathcal{N}_{2}$ in all. Node $\mathcal{N}_{1}$ and $\mathcal{N}_{2}$ have $k$ common neighbor nodes. We then calculate the NI between $\mathcal{N}_{1}$ and $\mathcal{N}_{2}$ as follows:
{\setlength\abovedisplayskip{2.5pt}
\setlength\belowdisplayskip{2pt}
\begin{equation} 
NI=Aggregation \left (\frac{k}{n_1},\frac{k}{n_2}  \right ).
\end{equation} }
Common aggregation operations include \emph{mean, minimum and maximum}. Comparison in Section~\ref{ablation} shows that the \emph{maximum} function has the best performance. Figure~\ref{NI} illustrates the NI on graph and shows the specific calculation. We further implement the above calculation into a matrix operation. Given adjacency matrix $A\in \mathbb{R}^{N\times N}$, the number of mutual neighbors of all node pairs is $\widetilde{A}=A \,A$, with each element $\widetilde{a}_{ij}$ in $\widetilde{A}$ denoting the number of mutual neighbours for $\mathcal{N}_{i}$ and $\mathcal{N}_{j}$. Then the NI is formulated as:
%\vspace{-2mm}
{\setlength\abovedisplayskip{3pt}
\setlength\belowdisplayskip{0pt}
\begin{equation} 
NI=max((\widetilde{A}^{T}\, sum_{0})^{T},\widetilde{A}\, sum_{1}),
\end{equation} }
{\setlength\abovedisplayskip{-1pt}
\setlength\belowdisplayskip{1pt}
\begin{equation} \nonumber
%sum_{0}={[{\sum_{j}a_{1j}}^{-1}, \cdots, {\sum_{j}a_{Nj}}^{-1}]}^{T},
sum_{0}=vec{({\sum_{j}a_{\cdot j}}^{-1})},sum_{1}=vec{({\sum_{i}a_{i \cdot}}^{-1})},
\end{equation}}
\vspace{-5mm}
%\begin{equation} \nonumber
%sum_{1}={[{\sum_{i}a_{i1}}^{-1},\cdots, {\sum_{i}a_{iN}}^{-1}]}^{T}.
%\end{equation} 

For inference, we use node intimacy to represent the edge score and remove those edges whose score is below $\tau_{2}$. After this step, we expect that most wrong connections have been removed. We can thus directly read the face clusters from the affinity graph.

%Given adjacency matrix $A\in \mathbb{R}^{N\times N}$, $\sum_{j}A_{ij}$ represents the total number of edges linked to node $\mathcal{N}_{i}$. The number of the common neighbors of $\mathcal{N}_{i}$ and $\mathcal{N}_{j}$ can be calculated as $[a_{i1},a_{i2},\cdots,a_{iN}]\,{[a_{1j},a_{2j},\cdots,a_{Nj}]}^{T}$. Therefore, we calculate the number of mutual neighbours of all node pairs as follows,

\vspace{2pt}
\noindent\textbf{Complexity Analysis.} During inference, the main computation lies in the GCN and the node intimacy. Both of these calculations are sparse matrix multiplication, thus the complexity is $\textit{O}\left ( \left | \mathscr{E}  \right | \right )$, where $ \mathscr{E}$ denotes the edges in the affinity graph. For a $K$NN affinity graph with $N$ nodes, we have $\left | \mathscr{E}  \right |  \leqslant \left | KN  \right | $, thus the complexity increases linearly as the number of nodes in the graph increases.

\section{Experiments}
\subsection{Experimental Settings}

\noindent\textbf{Datasets}.
We use MS1M~\cite{guo2016ms} and a large face benchmark named WebFace42M~\cite{FB} for training and testing in \emph{face clustering}. We follow the noisy list provided in~\cite{deng2019arcface} to clean the MS1M, and the refined MS1M contains about 5.82M images from 85K identities. We follow the setting in~\cite{yang2019learning} to partition MS1M~\cite{guo2016ms} into 10 splits with almost equal number of identities, while 1 part as labeled data $(\emph{part0\_ train})$ for training and the other 9 parts $( \emph{part1\_ test}, \cdots, \emph{part9\_ test} )$ as unlabeled data for testing. Each part consists of about 0.5M images from 8.6K identities. The WebFace42M is a new million-scale face benchmark including about 42M images and 2M identities which are cleaned from 260M images. It has near 7 times more images than MS1M thus presents a new challenge for face clustering. The MegaFace~\cite{kemelmacher2016megaface} is used to evaluate the \emph{face recognition} performance of the model trained using pseudo-labeled face images. It consists of a probe set with 3,530 images and a gallery set with over 1M images.

\begin{figure}[tb]
\begin{center}
\includegraphics[width=0.85\linewidth]{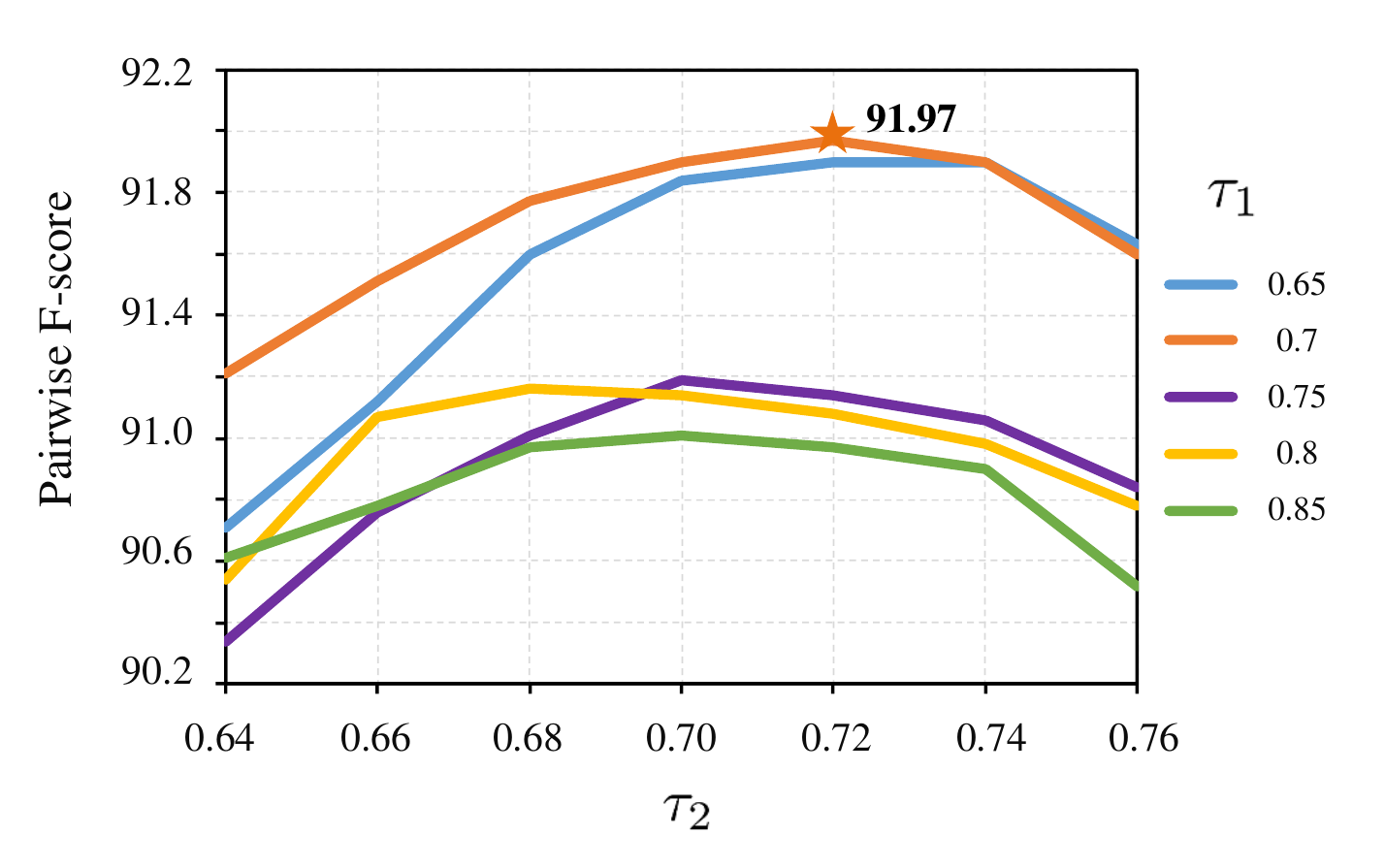}
\end{center}
\vspace{-8mm}
\caption{\emph{Pairwise F-score} under different selections of two thresholds $\tau_{1}$ and $\tau_{2}$.}
\vspace{-4mm}
\label{threshould}
\end{figure}

\begin{table}[tb]
\begin{center}
\vspace{3mm}
\centering
\begin{tabular}{c| c c c|c}
\hline
Method & Precision & Recall & $F_{P}$ & $F_{B}$\\
\hline\hline
Naive Pruning& 92.83 & 74.24&82.50&80.93\\
NI (mean) & 94.21 & 84.97 & 89.35&87.58\\
NI (min) &95.18  &81.93  &88.06&86.08\\
NI (max) & \textbf{95.50}&\textbf{85.91}&\textbf{90.45}&\textbf{88.06} \\
\hline
\end{tabular}
\end{center}
\vspace{-6mm}
\caption{Comparison between the naive and the NI-based pruning with different aggregation functions.}
\vspace{-5mm}
\label{table9}
\end{table}

\begin{comment}
\noindent\textbf{Training and Testing}.
We follow the setting in~\cite{yang2019learning} to partition MS1M~\cite{guo2016ms} into 10 splits with almost equal number of identities, while 1 part as labeled data $(\emph{part0\_ train})$ for training and the other 9 parts $( \emph{part1\_ test}, \cdots, \emph{part9\_ test} )$ as unlabeled data for testing. Each part consists about 0.5M images from 8.6K identities. For the WebFace42M~\cite{FB}, we randomly sample about 0.4M, 4M, 12M and 20M images respectively as labeled training data to verify the advantages of our approach in dealing with large-scale training sets. We then take 12M images as unlabeled data for testing. Particularly, the unlabeled data and the labeled set are orthogonal strictly. Most of the existing methods only select a portion of the MS1M~\cite{guo2016ms} for training and testing. However in practical application, very large scale training data is available and we need to deal with large scale face clustering inference. In this paper, we use the million-scale dataset WebFace42M~\cite{FB} to approximate these practical scenarios. 
\end{comment}

\vspace{3pt}
\noindent\textbf{Metrics}.
We evaluate the performance of our approach on both \emph{face clustering} and \emph{face recognition} tasks. For face clustering, we adopt the commonly used metric \emph{Pairwise F-score} ($F_{P}$) and \emph{BCubed F-score} ($F_{B}$)~\cite{amigo2009comparison}. For face recognition, we use different proportions of pseudo-label data along with 1 part labeled data to train the face recognition model and then test the rank-1 face identification accuracy on MegaFace challenge 1 with 1M distractors.

\noindent\textbf{Implementation Details}.
The affinity graph is built by $K$NN algorithm~\cite{cover1967nearest} with $K=80$ for MS1M~\cite{guo2016ms} and $K=30$ for WebFace42M~\cite{FB}. For structure-preserved subgraph sampling, we set $M=2$ (the number of cluster seeds), $N=750$ (the number of the sampled near clusters for each seed), $K_1=1300$ (parameter in CR) for MS1M and $M=4,N=1100,K_1=4000$ for WebFace42M, then set $K_2=90\%$ (parameter in SR) for both datasets.

\begin{table}[tb]
\begin{center}
\centering
\renewcommand\tabcolsep{3pt}
\begin{tabular}{c |c c c| c |c |c } 
\hline
 Method & SPSS & CR&SR& Nodes per batch&$F_{P}$&$F_{B}$ \\
\hline\hline
a& & & &${\sim 500K}$& 90.45&88.06\\
 b&\checkmark&&&${\sim 10K}$&91.21&89.19 \\
 c&\checkmark&\checkmark&&${\sim 10K}$&91.80&90.05 \\
 d&\checkmark&&\checkmark&${\sim 10K}$&91.92&90.06\\
 e&\checkmark&\checkmark&\checkmark&${\sim 10K}$&\textbf{91.97}&90.21 \\
\hline
\end{tabular}
\end{center}
\vspace{-6mm}
\caption{Comparison of different sampling strategies.}
\vspace{-5mm}
\label{table1}
\end{table}

\subsection{Ablation Study}
\label{ablation}
All models in this subsection are trained with the \emph{part0\_ train} and tested on the \emph{part1\_ test} in MS1M.

\begin{comment}
\noindent\textbf{Method Evolution}. In our approach, we design two modules, \ie GCN and node intimacy (NI), to gradually refine the initial affinity graph. Meanwhile, the struct-preserve subgraph sampling (SPS) strategy is proposed for more effective training. We can further add some randomness to this sampling strategy as follows. \emph{R1:} Randomly select a portion of clusters from the subgraph. \emph{R2:} Randomly select a portion of nodes from the subgraph. Table~\ref{table1} shows the gradual reinforcement of our approach with the above strategies. All models are trained with \emph{part0\_ train} and tested on \emph{part1\_ test} in MS1M. Experimental results show that using node intimacy for further edge pruning can boost the \emph{F-score} from 82.5 to 90.23. Adding the struct-preserve subgraph sampling (SPS) , the training speed will be accelerated, what's more, since such sampling strategy can enhance the generalization of the model, the face clustering performance is improved to 91.1. Further, adding some randomness to the SPS will bring extra performance gain. 
\end{comment}

\vspace{3pt}
\noindent\textbf{Selection of Thresholds}. In our method,  we refine the affinity graph with two steps involving two thresholds $\tau_{1}$ and $\tau_{2}$. In the first step, $\tau_{1}$ is used to process the GCN output edge scores, while in the second step $\tau_{2}$ is employed to prune the edges with low intimacy. We conduct experiments with different $\tau_{1}$, $\tau_{2}$. Results in Figure~\ref{threshould} show that $\tau_{1}=0.7$, $\tau_{2}=0.72$ is a suitable choice, we therefore adopt this setting in the following experiments.

\begin{table*}[tb]
\centering
\renewcommand\tabcolsep{10pt}
\begin{tabular}{l|cc|cc|cc|cc}
\hline
\#unlabeled & \multicolumn{2}{c|}{1.74M} & 
\multicolumn{2}{c|}{2.89M} & 
\multicolumn{2}{c|}{4.05M} &\multicolumn{2}{c}{5.21M} \\ \cline{1-9}
Method / Metrics  & $F_{P}$ & $F_{B}$  & $F_{P}$ & $F_{B}$ &$F_{P}$ & $F_{B}$ &$F_{P}$ & $F_{B}$ \\
\hline\hline
K-Means~\cite{lloyd1982least}&73.04& 75.20& 69.83& 72.34& 67.90& 70.57& 66.47& 69.42\\
HAC~\cite{sibson1973slink}&54.40& 69.53& 11.08& 68.62& 1.40& 67.69& 0.37& 66.96\\
DBSCAN~\cite{ester1996density}&63.41& 66.53& 52.50& 66.26& 45.24& 44.87& 44.94& 44.74\\
ARO~\cite{otto2017clustering}&8.78& 12.42& 7.30& 10.96& 6.86& 10.50& 6.35& 10.01\\
CDP~\cite{lloyd1982least} & 70.75&75.82&
69.51&74.58&68.62&73.62&68.06&72.92\\
L-GCN~\cite{wang2019linkage} &75.83&81.61&74.29&80.11&73.70&79.33&72.99&78.60\\
GCN-D~\cite{yang2019learning} &83.76&83.99&81.62&82.00&80.33&80.72&79.21&79.71\\
GCN-V+E~\cite{yang2020learning}&84.04&82.84&82.10&81.24&80.45&80.09&79.30&79.25\\
\hline\hline
\textbf{STAR-FC} &\textbf{88.28}&\textbf{86.26}&\textbf{86.17}&\textbf{84.13}&\textbf{84.70}&\textbf{82.63}&\textbf{83.46}&\textbf{81.47}\\
\hline
\end{tabular}
\vspace{-2mm}
\label{tab:exp_ms1m}
\caption{Comparison on face clustering when training with 0.5M face images and testing with different numbers of unlabeled face images. All results are obtained on the MS1M dataset. The proposed STAR-FC consistently outperforms other face clustering baselines on different scale of testing data.}
\vspace{-4mm}
\label{table3}
\end{table*}

\begin{table}[tb]
\begin{center}
\centering
\begin{tabular}{c c c c c}
\hline
Method & Precision & Recall & $F_{P}$ &Time\\
\hline\hline
K-Means~\cite{lloyd1982least} & 52.52 & 70.45 & 60.18 & 11.5h\\
DBSCAN~\cite{ester1996density} & 72.88 & 42.46 & 53.50 & \textbf{110s}\\
HAC~\cite{sibson1973slink} & 66.84 & 70.01 & 68.39 & 12.7h\\
ARO~\cite{otto2017clustering} & 81.10 & 7.30 & 13.34 & 1650s\\
CDP~\cite{zhan2018consensus} & 80.19 & 70.47 & 75.01 &  140s\\
L-GCN~\cite{wang2019linkage} & 74.38&83.51&78.68&5208s\\
GCN-D+S~\cite{yang2019learning} & \textbf{98.24} & 75.93 & 85.66 & 3700s\\
GCN-V+E~\cite{yang2020learning}& 92.56&83.74&87.93&690s\\
DA-Net~\cite{guo2020density} &95.88&85.87&90.60&329s\\
\hline\hline
\textbf{STAR-FC} & 96.20&\textbf{88.10}&\textbf{91.97}&310s\\
\hline
\end{tabular}
\end{center}
\vspace{-6mm}
\caption{Methods comparison on face clustering performance and inference time. All models are trained with part0\_train (0.5M images) from MS1M and tested with part1\_test (0.5M images) from MS1M. The STAR-FC significantly outperforms the state-of-the-arts, and can control the inference time within 310s.}
\vspace{-6mm}
\label{table2}
\end{table}

\begin{table*}[tb]
\centering
\renewcommand\tabcolsep{4.3pt}
\begin{tabular}{l|ccc|c|ccc|c|ccc|c|c}
\hline
\#unlabeled & \multicolumn{4}{c|}{4M} & 
\multicolumn{4}{c|}{8M} & 
\multicolumn{5}{c}{12M} \\ \cline{1-14}
Method / Metrics  & Pre & Recall & $F_{P}$&$F_{B}$ & Pre & Recall & $F_{P}$&$F_{B}$ &Pre & Recall & $F_{P}$&$F_{B}$&Time  \\
\hline\hline
K-Means~\cite{lloyd1982least} & 95.99&50.05&65.80&78.29& 92.20&49.91&64.34&76.47&88.75&49.69&63.71&75.04&2h\\
HAC~\cite{sibson1973slink}&98.25&59.76&74.31&85.46&96.55&58.98&73.23&84.57&OOM&OOM&OOM&OOM&OOM\\
DBSCAN~\cite{ester1996density}&94.77&44.12&60.21&77.87&89.97&43.55&58.69&77.02&85.57&43.76&57.91&76.38&3h \\
ARO~\cite{otto2017clustering} &\textbf{99.34}&62.83&76.98&88.83&\textbf{98.44}&62.01&76.09&88.66&\textbf{97.49}&62.34&76.05&88.60&4h \\
GCN-D~\cite{yang2019learning} &98.05&52.54&68.42&71.47&96.47&51.82&67.42&71.24&95.08&53.70&68.63&72.39&8h\\
\hline\hline
\textbf{STAR-FC} &96.77&\textbf{94.00}&\textbf{95.36}&\textbf{94.93}&93.95&\textbf{93.99}&\textbf{93.97}&\textbf{94.77}&90.86&\textbf{94.06}&\textbf{92.43}&\textbf{94.63}&\textbf{1.7h}\\
\hline
\end{tabular}
\label{tab:exp_ms1m}
\vspace{-2mm}
\caption{Comparison on face clustering when training with 4M face images and testing with different numbers of unlabeled data from the WebFace42M. Inference time on 12M testing data is shown in the right-most column. GCN-V+E~\cite{yang2020learning} fails to perform training on 4M data due to out-of-memory, so we don't show it in this table. HAC~\cite{sibson1973slink} is able to train on large-scale data, however it fails to perform large-scale inference with 12M testing data. The proposed STAR-FC achieves superior results on different testing settings and can complete inference on 12M data within 1.7h. }
\vspace{-3mm}
\label{table5}
\end{table*}
%irreplaceability
\vspace{3pt}
\noindent\textbf{Design of Node Intimacy}. In our method, we transform the face clustering into two steps: graph parsing with a trained GCN and graph refinement with node intimacy (NI). In this subsection, we investigate the impact of NI for the final face clustering performance, and compare three designs of NI. For the naive pruning method in Table~\ref{table9}, face clusters are obtained with dynamic edge pruning~\cite{zhan2018consensus} based on the edge scores predicted in the graph parsing step. Results in Table~\ref{table9} show that compared with the naive pruning strategy, using node intimacy for further graph refinement can significantly boost the \emph{pairwise F-score} from 82.5 to 90.45. This reveals the superiority of NI in dealing with face clustering problem. We further compare three different aggregation functions, \ie \emph{mean, minimum and maximum} for NI. Results in Table~\ref{table9} show that the \emph{maximum} strategy outperforms the other two methods. Therefore, we choose the \emph{maximum} strategy in the following experiments.

% To prove the irreplaceability of these two steps, we treat the second step as an ablation item and conduct two sets of experiments, \ie naive pruning and NI in Table~\ref{table9}. In naive pruning method, face clusters are obtained with dynamic edge pruning~\cite{zhan2018consensus} based on the edge scores predicted in the graph parsing step. Resluts in Table~\ref{table9} shows that the two-step refinement can significantly boost the pairwise F-score from 82.5 to 90.45, since the graph parsing step clears the graph structure and lays a good foundation for the implementation of the node intimacy. Thus these two steps are indispensable to each other. We further investigate three different aggregation functions, \ie mean, minimum and maximum for node intimacy. Results in Table~\ref{table9} shows that the maximum strategy outperforms the other two methods. We therefore choose it in the following experiments.

\vspace{3pt}
\noindent\textbf{Effect of Sampling Strategy}. In the training process, we propose the structure-preserved subgraph sampling (SPSS) strategy. To add more randomness, We further introduce \emph{cluster randomness} (CR) by randomly sampling partial clusters from the subgraph and \emph{sample randomness} (SR) by randomly sampling some nodes from the subgraph. In this subsection, we study the effect of SPSS for GCN training. As shown in Table~\ref{table1}, for the non-sampling method (a), it needs to take the whole graph with about 500K nodes per batch for training which leads to high GPU memory consumption. Nevertheless, equipped with the struct-preserved subgraph sampling strategy, our method only uses a subgraph with about 10K nodes per batch for training, and we achieve 91.21 \emph{pairwise F-score} which is comparable with the non-sampling method with less GPU memory usage. This interseting performance gain demonstrates that our sampling strategy successfully preserves most structure message of the whole graph. Moreover, adding the cluster and sample randomness to the SPSS can further improve the \emph{pairwise F-score} from 91.21 to 91.97. We argue that the introduction of randomness enhances the generalization of the trained model. These experimental results prove the ability of our method to handle large-scale training effectively. With the proposed SPSS, we can break through the limitation of the size of the training set and achieve excellent face clustering performance.

%No matter how large the original training set is, we can use a  for training and  at the same time. We can therefore break the limitation on the size of the input graph and explore larger-scale training.

%Training on sampled subgraphs, the GPU usage can be greatly reduced. The model can therefore handle larger-scale training set. Furthermore, since the subgraph sampling can enhance the generalization of the trained model, it brings extra performance gain. In Table~\ref{table1} we investigate the effects of different sampling strategies, \ie SPSS, R1 and R2. For specific description of these strategies, please refer to Algorithm~\ref{alg:sample}. The results show that after being equipped with the struct-preserve subgraph sampling, the pairwise F-score can be improved to 91.21. Moreover, adding some cluster and sample randomness to the SPSS can bring further improvement. 

\subsection{Face Clustering on MS1M}

Table~\ref{table3} and Table~\ref{table2} show the comparison on face clustering. All results in Table~\ref{table2} are obtained on the MS1M dataset with \emph{part0\_ train} as the training set and \emph{part1\_ test} as the testing set, and the inference time is obtained following the experimental configuration in ~\cite{yang2020learning}. In Table~\ref{table3} we further show the face clustering performance on different numbers of unlabeled data. Results in Table~\ref{table2} show that the proposed STAR-FC outperforms other clustering baselines consistently. Moreover, since all modules in the STAR-FC use full graph operation and parallel matrix computing, it can perform efficient inference within 310s. For fair comparison with DA-Net~\cite{guo2020density}, this time does not include the time of computing KNN graph, and the total inference time including computing KNN is 435s which can be accelerated with parallel GPUs. Results in Table~\ref{table3} show that our method can keep superior performance when dealing with larger-scale inference. What's more, compared with the representative clustering method GCN-V+E~\cite{yang2020learning}, our method boosts the $F_P$ significantly from 79.3 to 83.46 and improves the $F_B$ from 79.25 to 81.47.

\begin{figure}[tb]
\begin{center}
\vspace{-5mm}
\includegraphics[width=0.92\linewidth]{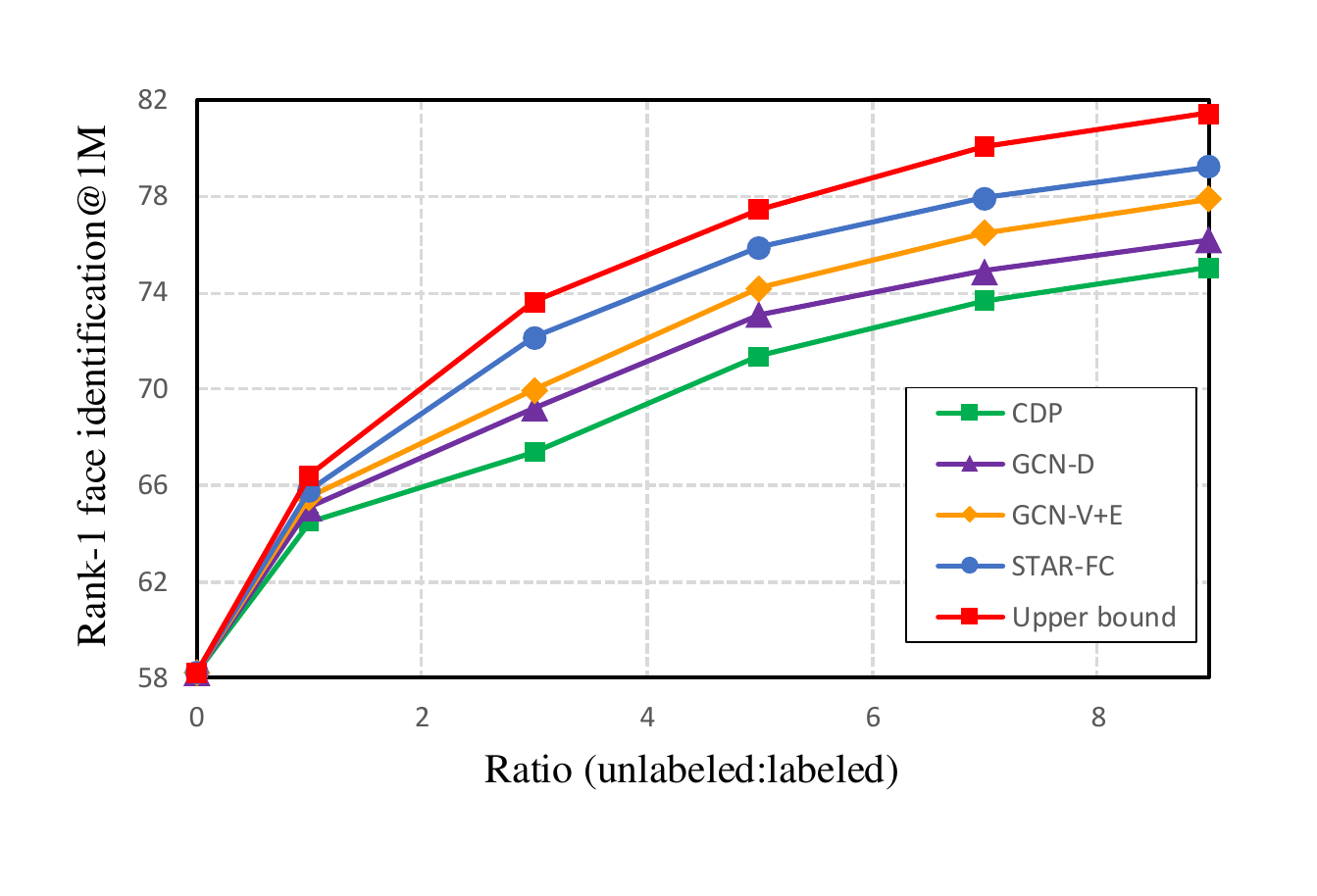}
\end{center}
\vspace{-12mm}
\caption{Rank-1 face identification accuracy on MegaFace with 1M distractors. The X-axis indicates the ratio of unlabeled to labeled data. The point where ratio is 0 indicates that only a split of labeled data is used for training. The upper bound is trained using data with ground-truth labels.}
\vspace{-5mm}
\label{fig5}
\end{figure}

We further use these pseudo-labeled data to train face recognition models and investigate the performance gain brought by these extra pseudo-labeled training data.  We follow the experiment setting in~\cite{yang2020learning,yang2019learning}, and use labeled data and various amounts of unlabeled data with pseudo-label to train the face recognition models. Figure~\ref{fig5} shows the rank-1 face identification accuracy on MegaFace~\cite{kemelmacher2016megaface} with 1M distractors. As shown in Figure~\ref{fig5}, extra unlabeled training data with pseudo-label brings continuous performance gain for face recognition. Owing to the superior performance in face clustering, our method achieves higher recognition accuracies than other face clustering baselines. With extra 5.21M unlabeled data, our method improves the recognition performance on MegaFace from 58.2\% to 79.26\%.

\subsection{Face Clustering on WebFace42M}
In this subsection, we first compare the face clustering performance of different methods on the WebFace42M and then explore the training upper bound of the STAR-FC.

Recent years have witnessed the success of large-scale training in many computer vision tasks. Large-scale training data is one of the key engines for the performance gain. To verify the capacity of the proposed method to handle large-scale graph, we conducte more experiments on the million-scale face benchmark WebFace42M~\cite{FB}. We randomly select 4M samples from the dataset as labeled data for training and take 4M, 8M and 12M samples respectively as unlabeled data for testing. There is no identities overlap between the training set and the testing set. We reproduce a series of clustering baselines on the WebFace42M dataset. Table~\ref{table5} shows their clustering performance and inference time on 12M testing data with acceleration by faiss~\cite{johnson2019billion}. Given the large-scale graph with 4M nodes for training, GCN-V will be directly out of memory. HAC fails to handle large-scale inference when the testing data size increases up to 12M. Under such settings of large-scale training and large-scale testing, our method once again achieves superior face clustering performance, and can complete the inference efficiently on 12M testing data within 1.7h. %What's more, we are the first to conduct face clustering training on a very large-scale graph with 4M nodes, thus provide a strong baseline for large-sclae face clustering. 

Furthermore, we make an exploration about the training upper bound of the proposed STAR-FC. We gradually increase the size of the training set and observe the performance changes. Specifically, we randomly select different scales of data (0.4M, 4M, 12M, 20M) from WebFace42M as labeled data for training, and test their face clustering performance on a testing set with 12M data. Experimental results in Table~\ref{table8} and Figure~\ref{fig1} show that global-based methods such as GCN-V fail to handle large-scale training while the local-based method GCN-D has poor performance. Nevertheless, with the increase of training data, our method has been consistently improved and finally achieves 95.1 \emph{pairwise F-score}. These experiments prove the performance superiority of the STAR-FC and its ability to handle large-scale training. What's more, we are the first to conduct face clustering training on a very large-scale graph with $10^{7}$ nodes, thus provide a strong baseline for large-scale face clustering. Our method is promising to perform excellently when a larger training set appears.

\begin{table}[tb]
\begin{center}
\centering
\renewcommand\tabcolsep{2.8pt}
\begin{tabular}{c |c c c|c c c| c}
\hline
Training set& Pre & Recall & $F_{P}$ &Pre & Recall & $F_{B}$ &NMI\\
\hline\hline
0.4M& 96.7&84.6 &90.2&99.6&75.9&86.1& 97.8\\
 4M& 90.9 & 94.1 &92.4&99.0&90.7&94.6&99.1\\
 12M& 95.4 & 92.9 & 94.2&99.4&88.3&93.5&99.0\\
 20M& 97.8 & 92.5 & 95.1&99.4&88.1&93.4&99.0\\
\hline
\end{tabular}
\end{center}
\vspace{-5mm}
\caption{Face clustering performance of the STAR-FC with different scales of training data from the WebFace42M and testing on 12M unlabeled data from the WebFace42M.}
\vspace{-5mm}
\label{table8}
\end{table}

\section{Conclusion}
In this paper, we have proposed a structure-aware face clustering method STAR-FC which addresses the dilemma of large-scale training and efficient inference. A structure-preserved subgraph sampling method is introduced to explore the power of larger-scale training data, and it can achieve satisfactory performance with less GPU memory usage. Moreover, a two-step graph refinement strategy with full-graph operation is developed to perform efficient inference. For the first time, a face clustering model is trained on a very large-scale graph with $10^{7}$ nodes. Extensive experiments on MS1M and WebFace42M demonstrate the superior face clustering performance of the proposed STAR-FC.

\vspace{7pt}
\noindent \textbf{Acknowledgement} This work was supported in part by the National Key Research and Development Program of China under Grant 2017YFA0700802, in part by the National Natural Science Foundation of China under Grant U1813218, Grant 61822603, Grant U1713214, in part by Beijing Academy of Artificial Intelligence (BAAI), and in part by a grant from the Institute for Guo Qiang, Tsinghua University.

{\small
\bibliographystyle{ieee_fullname}
\bibliography{egbib}
}

\clearpage
\appendix

\section{Effect of Graph Parsing}
\begin{table}[h]
\begin{center}
\centering
\renewcommand\tabcolsep{3.5pt}
\begin{tabular}{c c c c}
\hline
Method & Pre & Recall & $F_{P}$\\
\hline\hline
only graph refinement &91.84 & 70.87 & 80.00\\
only graph parsing& 92.83 & 74.24&82.50\\
graph parsing+graph refinemnet & \textbf{95.50}&\textbf{85.91}&\textbf{90.45} \\
\hline
\end{tabular}
\end{center}
\vspace{-6mm}
\caption{Method comparison with different inference strategies. Train with the \emph{part0\_ train} and test on the \emph{part1\_ test} in MS1M~\cite{guo2016ms}.}
\vspace{0mm}
\label{app_table1}
\end{table}

In the proposed STAR-FC, we transform the face clustering task into two steps: \textbf{graph parsing} and \textbf{graph refinement}. The graph parsing step is based on a GCN edge confidence estimator, while the graph refinement step is based on the node intimacy (NI). In our paper, we have proved that these two steps are indispensable. Here we conduct more experiments to further demonstrate that clustering faces with a single step does not work well.

In Table~\ref{app_table1} we compare the \emph{pairwise F-score} under three different inference strategies. The \emph{only graph refinement} method means that we directly perform pruning with NI on the \emph{K}NN graph. In \emph{only graph parsing} method, face clusters are obtained by dynamic edge pruning~\cite{zhan2018consensus} based on the GCN predicted edge scores. As shown in Table~\ref{table1}, face clustering with single step achieves poor performance. We have analyzed the disadvantage of the \emph{only graph parsing} method in the paper. Here we mainly discuss the poor performance of the \emph{only graph refinement} method. We argue that since there exist lots of wrong connections in the initial \emph{K}NN graph, different clusters are also densely linked and the NI of negetive neighbor may be very high. So the NI is difficult to work well on this graph. Therefore, performing graph parsing to get a relatively clear graph structure before employing the NI-based pruning is indispensable, and the combination of these two steps can achieve superior 90.45 \emph{pairwise F-score}.

\section{Comparing the NI with Jaccard similarity}
\begin{figure}[tb]
\begin{center}
\vspace{-4mm}
\includegraphics[width=0.6\linewidth]{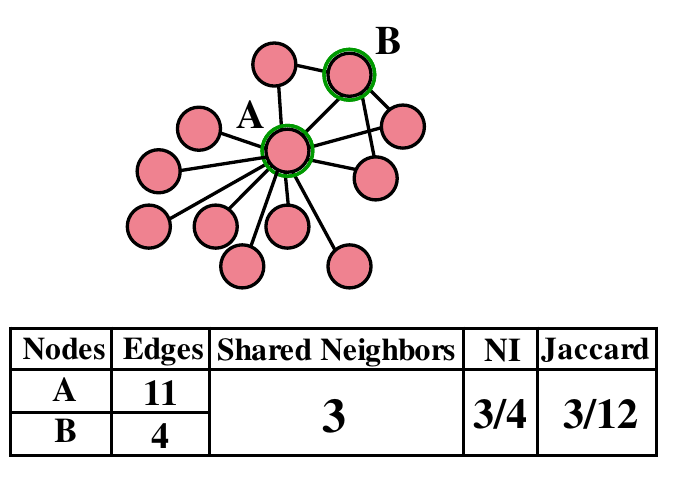}
\end{center}
\vspace{-6mm}
\caption{An example to visualize the difference of Jaccard similarity coefficient and the proposed NI.}
\label{app_fig1}
\end{figure}

Jaccard similarity coefficient is a statistic used for calculating the similarity and diversity of sample sets. Following the symbols defined in Sec.3.3, Jaccard~\cite{jaccard1912distribution,tanimoto1958elementary} $=\frac{k}{n_1+n_2-k}$, while NI $=max (\frac{k}{n_1},\frac{k}{n_2} )$. NI considers the attributes of two nodes respectively then uses max aggregation for  judgment while Jaccard ignores the difference between the two nodes. As shown in Figure~\ref{app_fig1}, \emph{A} and \emph{B} should be in the same cluster. Since \emph{A} has many neighbors and \emph{B} has a few, Jaccard gets a low score leading to misjudgment while NI can handle this case well. Therefore NI is more suitable for intimacy measures. In the added ablation study, the $F_{P}$ of Jaccard and NI are 88.39 and 91.97, respectively.

\begin{figure}[htb]
\begin{center}
\vspace{0mm}
\includegraphics[width=1.0\linewidth]{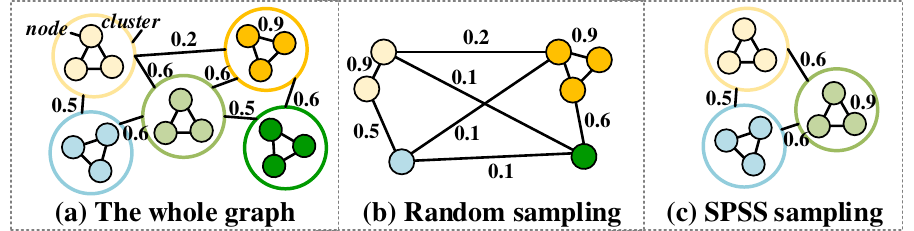}
\end{center}
\vspace{-5mm}
\caption{Comparison of random sampling method and the proposed SPSS sampling.}
\vspace{-2mm}
\label{app_fig3}
\end{figure}

\section{More details on the SPSS}

We use visualization to explain how SPSS works more vividly. Figure~\ref{app_fig3} illustrates the graph structure. (a) is the whole graph. (b) and (c) show the subgraph sampled in random way and SPSS respectively. Nodes with the same color belong to the same class. (c) preserves both the intra-cluster links and the \textbf{hard negative edges} between near clusters. \textbf{These two types of edges in (c) approximate the structure of edges in (a).} However the negative edges in (b) are mostly with low similarity that contribute less for training (Sec.3.2). Compared with the whole graph training (90.45 $F_{P}$), training with SPSS does not lead to performance loss and brings some extra accuracy gain (91.97 $F_{P}$), which further indicates that the global structure is preserved. 

\section{Discussion}
In this section, we discuss core differences between some representative face clustering methods and the proposed STAR-FC.

L-GCN~\cite{wang2019linkage} predicts the linkage within some selected subgraphs. It relies heavily on a mass of subgraphs. Since there exist many overlapped neighbors in these subgraphs, it suffers from heavily redundant calculations which is a big drag on the inference speed. Besides, such local graph operations lack the comprehension of global graph structure which limits its performance upper bound. GCN-D~\cite{yang2019learning} formulates face clustering as a detection and segmentation problem based on the affinity graph. However, it has the same problem as the L-GCN. It generates a large number of cluster proposals thus leading to inefficient inference. By contrast, the proposed STAR-FC performs face clustering inference based on the full-graph operation which can satisfy both efficiency and accuracy. ARO~\cite{otto2017clustering} computes the top-\emph{k} nearest neighbors for each face in the dataset and performs face clustering based on the approximate rank-order metric. However, it lacks the parsing of the initial structure information. The coarse top-\emph{k} nearest neighbors may contain a number of negative samples and the order of images may be far from the exact one, thus the effect of the rank-order metric will be greatly damaged. In the proposed STAR-FC, we use the edge scores predicted by the GCN to parse the graph in advance, therefore most wrong connections can be removed and the NI can work better on this graph with clearer structure. And the importance of graph parsing has been proved in Table~\ref{table1}. GCN-V+E~\cite{yang2020learning} obtains face clusters through predicting the vertex confidence and edge connectivity. It takes the entire graph as input for GCN training. Due to the limitation of GPU memory, it is hard to handle larger-scale training set. The proposed STAR-FC proposes the structure-preserved subgraph sampling strategy to address this challenge and is able to explore larger-scale training data.

\end{document}